# A Cluster-Cumulant Expansion at the Fixed Points of Belief Propagation

**Max Welling**
Dept. of Computer Science
University of California, Irvine
Irvine, CA 92697-3425, USA

**Andrew E. Gelfand**
Dept. of Computer Science
University of California, Irvine
Irvine, CA 92697-3425, USA

**Alexander Ihler**
Dept. of Computer Science
University of California, Irvine
Irvine, CA 92697-3425, USA

## Abstract

We introduce a new cluster-cumulant expansion (CCE) based on the fixed points of iterative belief propagation (IBP). This expansion is similar in spirit to the loop-series (LS) recently introduced in [1]. However, in contrast to the latter, the CCE enjoys the following important qualities: 1) it is defined for arbitrary state spaces 2) it is easily extended to fixed points of generalized belief propagation (GBP), 3) disconnected groups of variables will not contribute to the CCE and 4) the accuracy of the expansion empirically improves upon that of the LS. The CCE is based on the same Möbius transform as the Kikuchi approximation, but unlike GBP does not require storing the beliefs of the GBP-clusters nor does it suffer from convergence issues during belief updating.

## 1 Introduction

Graphical models play a central role in knowledge representation and reasoning across a broad spectrum of scientific disciplines, in which probabilistic queries, or *inference* must be performed. A classic goal is to calculate the *partition function*, or sum over all possible configurations of the model; this task is central to many problems, such as computing the probability of observed data ("probability of evidence"), learning from data, and model comparisons.

Exact inference in these systems is typically NP-hard, motivating the need for efficient and accurate approximations. One of the most successful algorithms is *iterative belief propagation* (IBP) [7], which approximates the marginal probabilities of the distribution via an iterative, message-passing procedure. IBP can also be interpreted as a variational optimization, in which the messages and their approximate marginals, or beliefs, minimize a particular free energy approximation called the Bethe approximation [11, 3]. IBP is approximate on graphs with cycles, but is often empirically very accurate.

The wide success of IBP has also led to several techniques for improving its estimate quality. Generalized belief propagation (GBP) performs IBP-like updates on a set of *regions*, consisting of collections of variables. However, region selection is often difficult, and can even lead to convergence problems and degraded estimates [10]. An alternative is to use "series corrections" to the partition function such as the loop series and its generalizations, which compute modifications to the estimate provided by a given fixed point of IBP [1]. Such series expansions can be used to provide "any-time" algorithms that initialize to the IBP estimate and eventually converge to the exact solution.

In this paper we propose an alternative series correction to IBP estimates based on a cluster cumulant expansion (CCE). We show that the CCE has several benefits over the loop series representation, including more accurate partial estimates and more efficient aggregation of terms. We also show that our CCE representation is closely related to GBP, and that it can naturally be extended to GBP fixed points. We show the effectiveness of our approach empirically on several classes of graphical models.

## 2 Background

Let $X = \{X_i\}$ be a collection of random variables, each of which takes on values in a finite alphabet, $X_i = x_i \in \mathcal{X}$. For convenience, in the sequel we will not distinguish between the random variable $X_i$ and its instantiation $x_i$. Suppose that the distribution on $X$ factors into a product of real-valued, positive functions $\{\psi_f : f \in F\}$, each defined over a subset $f$ of the variables:

$$p(\mathbf{x}) = \frac{1}{Z_\psi} \psi(\mathbf{x}) = \frac{1}{Z_\psi} \prod_{f \in F} \psi_f(x_f)$$

where $x_f = \{x_i \,|\, i \in f\}$ represents the arguments of factor $\psi_f$, and $Z_\psi$ is the partition function, $Z_\psi = \sum_x \psi(\mathbf{x})$, which serves to normalize the distribution. We denote by $F_i$ the set of factors that have $x_i$ in their argument. A *factor graph* represents this factorization using a bipartite graph, in which each factor (represented by a square) is connected to the variables (circles) in its argument. See Figure 1-Left.

## 2.1 Iterative Belief Propagation

With each factor or variable node in the factor graph (FG) we will associate a belief, denoted with $b_f(x_f)$ and $b_i(x_i)$ respectively. Iterative (or Loopy) Belief Propagation (IBP) is a message passing algorithm on the factor graph that attempts to make these beliefs consistent with each other. In particular, at a fixed point of IBP we require

$$\sum_{x_f \setminus x_i} b_f(x_f) = b_i(x_i) \qquad (1)$$

for every pair of factor and variables nodes that are connected in the factor graph. IBP expresses these beliefs in terms of a set of messages $m_{fi}$:

$$b_i(x_i) \propto \prod_{f \in F_i} m_{fi}(x_i)$$
$$b_f(x_f) \propto \psi_f(x_f) \prod_{i \in f} \prod_{g \in F_i \setminus f} m_{gi}(x_i) \qquad (2)$$

and updates the messages iterative to achieve Eqn. (1):

$$m_{fi}^{\text{new}}(x_i) \leftarrow \delta(x_i) m_{fi}^{\text{old}}(x_i), \quad \delta(x_i) \doteq \frac{\sum_{x_f \setminus x_i} b_f(x_f)}{b_i(x_i)}$$

If the factor graph contains no cycles, at convergence the beliefs exactly equal the marginal probabilities of $p(\mathbf{x})$.

[11] showed that IBP fixed points correspond to stationary points of the Bethe variational free energy $\mathcal{F}$, giving an approximation to the log-partition function:

$$\log \hat{Z}_{BP} = -\mathcal{F}(\{\psi_f\}, \{b_f, b_i\}) \qquad (3)$$
$$= \sum_{f \in F} \mathbb{E}_{b_f}[\log \psi_f] + \sum_{f \in F} \mathbb{H}(b_f) + \sum_{i \in V}(1 - |F_i|) \mathbb{H}(b_i)$$

where $V$ is the set of variables, $\mathbb{E}_{b_f}$ denotes the expectation under $b_f$ and $\mathbb{H}(b_f) = -\sum_x b(x_f) \log b(x_f)$ is its entropy.

We may view the IBP updates as a *reparameterization* of the original distribution [9]. In particular, if we apply the message update from factor $f$ to variable $i$, according to Eqn. (2) we change $b_i(x_i) \leftarrow b_i(x_i)\delta(x_i)$ and also a total of $|F_i|-1$ factor beliefs $b_g(x_g) \leftarrow b_g(x_g)\delta(x_i)$ with $g \in F_i \setminus f$ (see Figure 1-Left). As a result, the expression

$$\frac{1}{Z_b} b(\mathbf{x}) = \frac{1}{Z_b} \prod_{f \in F} b_f(x_f) \prod_{i \in V} b_i(x_i)^{1-|F_i|} \qquad (4)$$

remains invariant under message updating, where $Z_b$ is the partition function of the reparameterization $b(\mathbf{x})$. Moreover, if we initialize all messages to 1, we immediately see that Eqn. (4) is also equal to $\frac{1}{Z_\psi} \psi(\mathbf{x})$.

Since $\psi(\mathbf{x})$ and $b(\mathbf{x})$ correspond to the same normalized distribution $p(\mathbf{x})$, they differ by a constant multiplicative factor. Using (3) and noting that $\mathcal{F}(\{b\}, \{b_f, b_i\}) = 0$ shows that this factor is exactly $\hat{Z}_{BP}$, and so

$$Z_\psi = \hat{Z}_{BP} \cdot Z_b \qquad (5)$$

This shows that the true partition function $Z_\psi$ can be computed from IBP's estimate $\hat{Z}_{BP}$ and the reparameterization's normalization constant, $Z_b$.

## 2.2 Generalized Belief Propagation

Generalized BP, or GBP [11], generalizes (3) to include higher-order interactions than the original factors of the model. In GBP, one identifies a set of *regions* $\mathcal{R}$, in which each region $\alpha \in \mathcal{R}$ is defined to be a subset of the factors, $\alpha \subseteq F$. Each region is also given a "counting number" $c_\alpha$, and messages are passed between regions; a data structure called a *region graph* is used to organize the message passing process. The GBP estimate of the partition function is

$$\log \hat{Z}_{GBP} = \sum_{f \in F} \mathbb{E}_b[\log \psi_f] + \sum_{\alpha \in \mathcal{R}} c_\alpha \mathbb{H}(b_\alpha). \qquad (6)$$

BP on the factor graph constitutes a special case of GBP, in which the regions $\mathcal{R} = F \cup V$ and counting numbers $c_f = 1$ and $c_i = 1 - |F_i|$. Regions form a partial ordering defined by their set inclusion; the *ancestors* of region $\alpha$ are $an(\alpha) = \{\gamma \in \mathcal{R} | x_\alpha \subset x_\gamma\}$ and its *descendants* are $de(\alpha) = \{\beta \in \mathcal{R} | x_\alpha \supset x_\beta\}$, where $x_\alpha$ is the set of variables in region $\alpha$.

The counting numbers should satisfy some basic properties; in [11], a RG is considered *valid* or *1-balanced* if

$$\sum_{\alpha \in R(f)} c_\alpha = 1 \,\forall\, f \qquad \sum_{\alpha \in R(i)} c_\alpha = 1 \,\forall\, i$$

where $R(f)$, $R(i)$ are those regions that contain factor $f$ and variable $i$, respectively.

Validity can be ensured by assigning each factor $f$ to a single outer region, setting $c_\alpha = 1$ for outer regions (i.e. regions with no parents) and recursively for inner regions as:

$$c_\alpha = 1 - \sum_{\beta \in an(\alpha)} c_\beta. \qquad (7)$$

The GBP message updates can again be interpreted as reparameterization updates on the region graph. Any two regions that are connected through a directed edge will exchange messages from parent to child region. We write the beliefs in terms of the potentials and messages to $\alpha$ and its descendants, $\Delta_\alpha = \alpha \cup de(\alpha)$,

$$b_\alpha(x_\alpha) = \frac{1}{Z_\alpha} \prod_{f \in \alpha} \psi_f(x_f) \prod_{\substack{\gamma \in \{an(\Delta_\alpha) \setminus \Delta_\alpha\} \\ \beta \in \Delta_\alpha}} m_{\gamma\beta}(x_\beta)$$

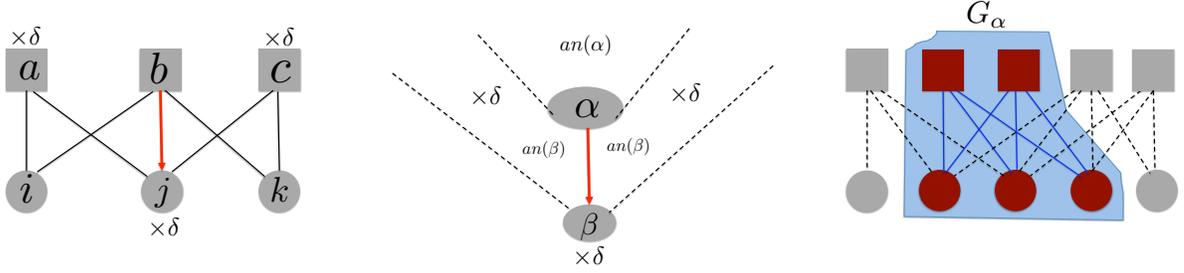

Figure 1: **Left:** A reparameterization update on a factor graph. **Middle:** A region graph GBP reparameterization update. **Right:** Factor subgraph $G_\alpha$ corresponding to a cluster-cumulant.

and write a fixed-point message update as,

$$m_{\alpha\beta}^{\text{new}}(x_\beta) \leftarrow \delta(x_\beta) m_{\alpha\beta}^{\text{old}}(x_\beta) = \frac{\sum_{x_\alpha \setminus x_\beta} b_\alpha(x_\alpha)}{b_\beta(x_\beta)} m_{\alpha\beta}^{\text{old}}(x_\beta)$$

In words, to compute the belief at region $\alpha$ we collect messages from all regions that are either ancestors or ancestors of descendants of that region, and that flow into $\alpha$ or its descendants. Those messages that originate in $\alpha$ or its descendants are not used.

It is now not difficult to see that these updates constitute a reparameterization. Namely, if we update a message $m_{\alpha\beta}$, the update factor $\delta$ is multiplied into the expressions for $b_\beta$ and all its ancestors, except $b_\alpha$ and all its ancestors. Using the expression for the counting numbers (7) we thus see that we accumulate a total number of terms,

$$c_{\text{total}} = [c_\beta + \sum_{\gamma \in an(\beta)} c_\gamma] - [c_\alpha + \sum_{\delta \in an(\alpha)} c_\delta] = 0$$

where we used that $c_\beta = 1 - \sum_{\gamma' \in an(\beta)} c_{\gamma'}$ and similarly for $c_\alpha$. This is illustrated in Figure 1-Middle.

As a result, the following expression will stay invariant under the GBP updates:

$$\frac{1}{Z_\psi} \prod_f \psi_f(x_f) = \frac{1}{Z_b} \prod_\alpha b_\alpha(x_\alpha)^{c_\alpha} \qquad (8)$$

We can now use similar arguments as in Section 2.1 to show that $Z_\psi = \hat{Z}_{GBP} \cdot Z_b$ again holds.

### 2.3 The Loop Series Expansion

The loop series [1, 8, 2] expresses $Z_b$ as a finite series expansion of terms related to the generalized loops of the graph. Starting from Eqn. (4) we rewrite the terms:

$$\frac{b_f(x_f)}{\prod_{i \in f} b_i(x_i)} = 1 + \frac{b_f(x_f) - \prod_{i \in f} b_i(x_i)}{\prod_{i \in f} b_i(x_i)} = 1 + U_f(x_f)$$

where $U_f$ is like a covariance in that it measures dependence among the variables $x_i, i \in f$. We can then write

$$Z_b = \mathbb{E}_{\tilde{b}}\Big[\prod_{f \in F}(1 + U_f)\Big] \quad \text{where} \quad \tilde{b}(\mathbf{x}) = \prod_i b_i(x_i).$$

Expanding the product gives one term for each element in the power set of $F$. If we expect the individual terms $U_f$ to be small, it makes sense to organize this series (for example) in order of increasing number of $U_f$ terms, and truncate it once some computational limit is reached. Each term corresponds to a subset $\alpha \subseteq F$, and thus to a subgraph of the factor graph. One can show that if this subgraph contains any factor $f$ that is *not* part of a loop (a.k.a. part of a dangling tree), then the corresponding term will be zero.

To make the connection with the formulation of [1], we must assume a binary alphabet $x_i \in (0, 1)$. Following [8] we can re-express,

$$\frac{b_f(x_f) - \prod_{i \in V_f} b_i(x_i)}{\prod_{i \in V_f} b_i(x_i)} = \sum_{v \subseteq V_f, |v| \geq 2} \beta_v \prod_{i \in v}(x_i - \mathbb{E}[x_i])$$

$$\beta_v = \frac{\mathbb{E}[\prod_{i \in v}(x_i - \mathbb{E}[x_i])]}{\prod_{j \in v} \mathbb{V}(x_j)} \qquad (9)$$

where $\mathbb{V}(x)$ denotes the variance of $x$. In the process we have created even more terms than the powerset of all factors because each factor itself is now a sum over all its subsets of variables with cardinality larger than one. The total set of terms is in one-to-one correspondence with all the "generalized loops" of the factor graph (graphs where every variable and factor node have degree at least two). The advantage for the binary case is that the final expansion can now be written in terms of expectations of the form $\mathbb{E}[(x_i - \mathbb{E}[x_i])^d]$ which admit closed form expressions in terms of the beliefs $b_i$ obtained from IBP at its fixed point. We refer to [1, 8] for further details.

## 3 The Cluster-Cumulant Expansion

We now develop an alternative expansion of the log-partition function which we call the cluster-cumulant expansion (CCE). As with the loop series, we begin analysis at a fixed point of IBP; we will later relate our expansion to GBP and extend its definition to include GBP fixed points.

The CCE is defined over an arbitrary collection of subsets of factors which we will denote with $\alpha$. We define a partial ordering on this set through subset inclusion, i.e. $\alpha \leq \beta$ iff $\alpha \subseteq \beta$ and $\alpha < \beta$ iff $\alpha \subset \beta$. We will denote this poset with $\Omega$. We also define a factor subgraph, $G_\alpha = \{i, f | i \in V_\alpha, f \in \alpha, \alpha \in \Omega\}$ where $V_\alpha$ corresponds to the set for variables that occur in the arguments of the factors $f \in \alpha$. An example of such a factor subgraph is provided in Figure 1-Right. Finally we will define the partial probability distribution over this factor subgraph as $P_\alpha$ and the corresponding partial log partition function, $\log Z_\alpha$ as its log normalization constant:

$$P_\alpha(x_\alpha) = \frac{1}{Z_\alpha} \prod_{i \in V_\alpha} b_i(x_i) \prod_{f \in \alpha} \frac{b_f(x_f)}{\prod_{i \in V_f} b_i(x_i)} \quad (10)$$

$$\log Z_\alpha = \log \sum_{x_\alpha} \prod_{i \in V_\alpha} b_i(x_i) \prod_{f \in \alpha} \frac{b_f(x_f)}{\prod_{i \in V_f} b_i(x_i)} \quad (11)$$

We first observe that if we choose $\alpha = F$ (the set of all factors) then $\log Z_F$ is the exact log partition function $\log Z_b$. We now show that $\log Z_\alpha = 0$ when the corresponding factor subgraph $G_\alpha$ is a tree:

THEOREM 1. *If the factor subgraph $G_\alpha$ is* singly connected, *then $\log Z_\alpha = 0$.*

*Proof.* When $G_\alpha$ is a tree, the expression (10) represents the exact joint probability distribution expressed in terms of its marginals with $Z_\alpha = 1$. Thus the result follows. □

As a corollary we observe that $\log Z_\alpha$ vanishes if $\alpha$ consists of a single factor $f$. The theorem also allows us to remove all "dangling trees" from any factor subgraph by marginalizing out the variables that correspond to the dangling tree. We will thus define the "core" of a factor subgraph as the part of the factor subgraph that remains after all dangling trees have been removed (see also [2]).

We are now ready to define the cluster-cumulants (CCs). The idea is to decompose each partial log partition function as a sum of cluster-cumulant contributions from clusters in the same or lower levels of the partially ordered set $\alpha \in \Omega$. In this manner, we expect that the lowest order CCs generate the largest contribution and that higher order CCs subsequently represent increasingly small corrections. The definition of the CCs is provided by the expression [5, 4]

$$\log Z_\alpha = \sum_{\beta \leq \alpha} C_\beta \quad (12)$$

This relation can be inverted using a Möbius transform,

$$C_\alpha = \sum_{\beta \leq \alpha} \mu_{\beta,\alpha} \log Z_\beta \quad (13)$$

where we define the Möbius numbers as,

$$\mu_{\alpha,\alpha} = 1 \quad \text{and} \quad \mu_{\gamma,\alpha} = -\sum_{\beta: \gamma < \beta \leq \alpha} \mu_{\beta,\alpha} \quad \text{if } \gamma < \alpha.$$

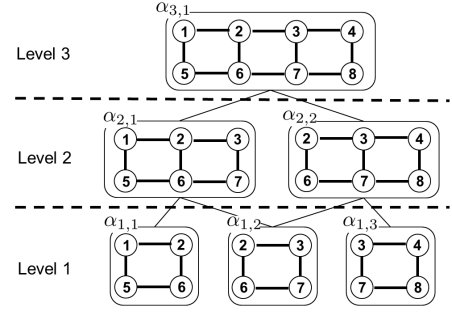

Figure 2: A poset for a pairwise MRF on a $1 \times 3$ grid. Truncating this poset at level 1 would include clusters over the faces of the grid. Truncating at level 2 would add clusters $\alpha_{2,1}$ and $\alpha_{2,2}$.

To approximate $\log Z \approx \log Z_\ell$ we truncate the cumulant series by only including clusters $\alpha \in \Omega_\ell$ up to some level $\ell$ in the poset:

$$\log Z_\ell = \sum_{\alpha \in \Omega_\ell} C_\alpha = \sum_{\alpha \in \Omega_\ell} \sum_{\beta \leq \alpha} \mu_{\beta,\alpha} \log Z_\beta \quad (14)$$

$$= \sum_{\beta \in \Omega_\ell} \sum_{\alpha \in \Omega_\ell} \mu_{\beta,\alpha} \mathbb{I}[\beta \leq \alpha] \log Z_\beta \doteq \sum_{\beta \in \Omega_\ell} \kappa_\beta \log Z_\beta$$

where $\mathbb{I}[\cdot]$ is the indicator function. Moreover, $\kappa_\beta \doteq \sum_{\alpha \in \Omega_\ell} \mu_{\beta,\alpha} \mathbb{I}[\beta \leq \alpha]$ can be computed recursively as

$$\kappa_\beta = 1 - \sum_{\alpha \in an(\beta)} \kappa_\alpha \quad (15)$$

with $an(\beta)$ the ancestors of cluster $\beta$ in the Hasse diagram corresponding to the poset and with $\kappa_\alpha = 1$ for the clusters at the highest level of $\Omega_\ell$ (i.e. the clusters with no parents) [6]. It is important to realize that this truncation is *not* equivalent to simply including all partial log partitions from the the highest level in the poset, i.e. $\log Z_\ell = \sum_{\alpha \in \Omega_\ell} C_\alpha \neq \sum_{\alpha \in \Omega_\ell} \log Z_\alpha$.

Figure 2 illustrates the computation of $\log Z_\ell$ on a simple pairwise MRF. On this $1 \times 3$ grid, the approximation truncated at level 2 is $\log Z_2 = \log Z_{\alpha_{2,1}} + \log Z_{\alpha_{2,2}} - \log Z_{\alpha_{1,2}}$. The computation of each partial log partition function requires exact inference on the partial factor subgraph. Thus, the time complexity for an approximation including $c$ clusters is $O(c|V| \exp(w))$, where $w$ is the largest induced width of the $c$ clusters. However, the space complexity of the truncated approximation is $O(|V| \exp(w))$ since we need only retain $\log Z_\alpha$ for each cluster.

This expansion is similar to the expansion of the free energy (or the entropy) in the cluster variation method of GBP (see Section 2.2). However, there clusters serve as regions between which messages are exchanged during the execution of GBP. In contrast, in this paper CCs are calculated after IBP has converged in order to compute corrections.

The cumulant definition in Eqn. (13) immediately shows that cumulants of singly connected subgraphs vanish:

THEOREM 2. *If the factor subgraph $G_\alpha$ is singly connected, then the cluster-cumulant satisfies $C_\alpha = 0$.*

*Proof.* From Eqn. (13) we see that a cumulant can be written as a linear combination of $\log Z_\alpha$ and $\log Z_\beta$, $\beta < \alpha$. Using Theorem 1 together with the facts that every subgraph of a tree is a smaller tree (or a forest of trees) and that at the lowest level of the poset we have $C_\alpha = \log Z_\alpha$, the result follows by induction. □

Unlike the terms in the loop series [1] (see Section 2.3), cluster-cumulants corresponding to *disconnected* subgraphs also vanish, even if the disconnected components are not singly connected.

THEOREM 3. *If the factor subgraph $G_\alpha$ is disconnected, then the cluster-cumulant satisfies $C_\alpha = 0$.*

*Proof.* If the graph $G_\alpha$ is disconnected, its probability $P_\alpha$ factorizes: $P_\alpha = P_{\alpha_A} P_{\alpha_B}$. Then, the partial partition function decomposes as $\log Z_\alpha = \log Z_{\alpha_A} + \log Z_{\alpha_B}$ which by Eqn. (12) we rewrite as $\log Z_\alpha = \sum_{\beta_A \leq \alpha_A} C_{\beta_A} + \sum_{\beta_B \leq \alpha_B} C_{\beta_B}$. Again by Eqn. (12) we have

$$C_\alpha = \log Z_\alpha - \sum_{\beta < \alpha} C_\beta$$
$$= \sum_{\beta_A \leq \alpha_A} C_{\beta_A} + \sum_{\beta_B \leq \alpha_B} C_{\beta_B} - \sum_{\beta < \alpha} C_\beta.$$

Clusters $\beta < \alpha$ in the poset fall into one of three categories: either $C_\beta \leq C_{\alpha_A}$, or $C_\beta \leq C_{\alpha_B}$, or $[(C_\beta > C_{\alpha_A}) \wedge (C_\beta > C_{\alpha_B}) \wedge (C_\beta < C_\alpha)]$. We now proceed by induction. Cumulants in the first two categories will cancel in the expression for $C_\alpha$. Sufficiently small clusters will have no cumulants in the third category, and so must be zero. For larger clusters, cumulants in the third category correspond to smaller clusters that must also be disconnected, which by the inductive argument must vanish. The result follows. □

This property is special to the cumulant expansion and holds even though the partial log-partition function $\log Z_\alpha$ for the disconnected region does *not* vanish. Apart from reducing the number of terms that must be considered, this property also suggests (by continuity of the cumulants as a function of their factor parameters) that cumulants with nearly-independent components must also be small. Thus, we expect significant contributions from tight, highly correlated sets of variables and factors, but small contributions from clusters with components that are almost independent.

To emphasize this point, consider a cluster-cumulant corresponding to a subgraph $G_\alpha$ that has one uniform factor $\psi_f(x_f) = 1$. Moreover, assume that the cluster $\tilde{\alpha}$ given by removing this factor from $\alpha$ is also in the poset. This creates a situation where cluster $\alpha$ is ranked higher in the poset than $\tilde{\alpha}$, i.e., $\alpha > \tilde{\alpha}$, yet their partial partition functions must be the same: $\log Z_\alpha = \log Z_{\tilde{\alpha}}$. In this case, $C_\alpha = 0$:

THEOREM 4. *Consider a poset $\Omega$ that includes a factor $f$ with unit potential $\psi_f(x_f) = 1$, and a cluster $\alpha$ that contains factor $f$. Assume that there is also a cluster $\tilde{\alpha} < \alpha$ containing the same factors as $\alpha$ except $f$, i.e. $\tilde{\alpha} = \alpha \backslash f$. Then, $C_\alpha = 0$.*

*Proof.* For both clusters $\{\alpha, \tilde{\alpha} | \alpha > \tilde{\alpha}\}$ we must have the same partial partition function, $\log Z_\alpha = \log Z_{\tilde{\alpha}}$. Hence, from Eqn. (12) we have $\sum_{\beta \leq \alpha} C_\beta = \sum_{\tilde{\beta} \leq \tilde{\alpha}} C_{\tilde{\beta}}$. Since $\alpha > \tilde{\alpha}$ it follows that $\sum_{\beta \leq \alpha} C_\beta = \sum_{\tilde{\beta} \leq \tilde{\alpha}} C_{\tilde{\beta}} + \sum_{\gamma > \tilde{\alpha}, \gamma \leq \alpha} C_\gamma$. However, since $\tilde{\alpha}$ is defined to have exactly one factor fewer than $\alpha$ we have that $\sum_{\gamma > \tilde{\alpha}, \gamma \leq \alpha} C_\gamma = C_\alpha$. Combining these expressions we have $C_\alpha = 0$. □

The significance of Theorem 4 is illustrated by imagining a situation where $\psi_f(x_f) \approx 1$. By the continuity argument, we expect $C_\alpha \approx 0$ if the cluster $\tilde{\alpha}$ with factor $f$ removed is in the poset. Hence, a cluster-cumulant will only significantly differ from zero if it introduces new dependencies that were not already captured by lower order cumulants. This result strengthens the interpretation of CCE as an expansion in terms of orders of statistical dependency.

These considerations suggest a way to build posets that are likely to deliver good truncated series approximations of the log-partition function. We first choose a collection of clusters which may be overlapping, but none are a subset of any other cluster. These clusters should be chosen so that 1) they are not disconnected and 2) they have no dangling trees. Moreover, tightly coupled clusters with strong interdependencies are preferred over ones with weak dependence. We then generate all intersections, intersections of intersections, etc., to construct our poset. Finally we compute their partial log partition functions and combine them using Eqn. (14).

## 4 CCE for Region Graphs

We now extend the cluster-cumulant expansion to region graphs; see Section 2.2 and [11] for background on region graphs. The CCE will be build on top of an existing region graph. We will call the existing region graph "calibrated" (abbreviated CRG) if the GBP fixed point equations have converged.[1] We will add new "uncalibrated" regions to the region graph during the expansion (see Figure 3-Left). Any new region will become a parent of all the existing regions that it contains. We then define the region subgraph $G_\alpha = G_{\Delta_\alpha}$ to be the collection of *calibrated* regions and parent-child relations restricted to $\alpha$ and its descendants. Note that $G_\alpha$ includes only regions in the calibrated part of the region subgraph, i.e., below the dashed line in Figure 3-Left.

---

[1] We only consider CCEs on calibrated region graphs; although an expansion can be established for uncalibrated RGs, Theorem 2 will not hold (increasing the number of non-zero cumulants) and terms may not fall off quickly, leading to poor approximations.

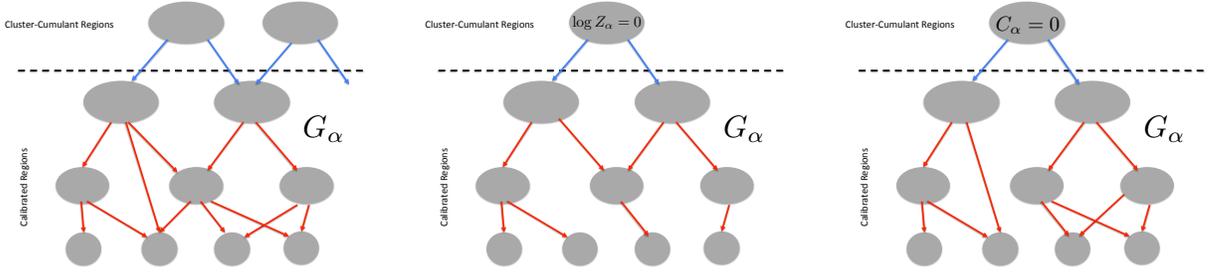

Figure 3: **Left:** Cluster-cumulant extension of the calibrated region graph. **Middle:** A tree-structured RG for which $\log Z_\alpha = 0$. **Right:** Illustration of a disconnected RG for which $C_\alpha = 0$.

We define counting numbers $c_\beta^\alpha$ for regions $\beta$ in $G_\alpha$ by

$$c_\beta^\alpha = 1 - \sum_{\gamma \in (G_\alpha \cap an(\beta))} c_\gamma^\alpha$$

Note that a counting number $c_\beta^\alpha$ can vanish for some regions in $G_\alpha$ even though their counting number $c_\beta$ in the original region graph did not vanish. Regions with $c_\beta^\alpha = 0$ will not contribute to $\log Z_\alpha$ and can be excluded from $G_\alpha$.

The partial log-partition function for this region is,

$$\log Z_\alpha = \log \sum_{x_\alpha} \prod_{\beta \subseteq \alpha} b_\beta(x_\beta)^{c_\beta^\alpha} \qquad (16)$$

To generate cumulants we again use Eqn. (13). The truncated cumulant series is the sum of these cumulants up to some level in the poset. More convieniently, we can instead use Eqn. (14) to express the series in terms of partial log partition functions. Note that this will require counting numbers $\kappa_\alpha$ that should not be confused with the counting numbers $c_\alpha$ above. In particular, the $c_\alpha$ are computed strictly in terms of the calibrated part of the region graph (e.g., the regions right below the dashed line in Figure 3-Left have $c_\alpha = 1$), while the $\kappa_\alpha$ are computed strictly in terms of the cluster-cumulants (e.g., the regions at the highest level in the extended region graph have $\kappa_\alpha = 1$).

As with the IBP expansion, $\log Z_\alpha$ vanishes for a region subgraph $G_\alpha$ that is singly connected (after only regions with $c_\beta^\alpha \neq 0$ have been retained):

THEOREM 5. *If a region subgraph $G_\alpha$ consisting of regions in the CRG that are descendants of $\alpha$ and that have counting numbers $c_\beta^\alpha \neq 0$ is singly connected, then the partial log partition function and the cumulant satisfy $\log Z_\alpha = 0$, $C_\alpha = 0$.*

*Proof.* Since the marginals on $G_\alpha$ are calibrated (due to the reparameterization property of GBP) and because $G_\alpha$ is singly connected, we have $Z_\alpha = 1$. Since according to Eqn. (13) cumulants are linear combinations of log-partition functions over subsets and since all subsets must also be singly connected (or disconnected), the cumulant for a singly connected region must also vanish. □

This idea is illustrated in Figure 3-Middle. More generally, for any region subgraph we can remove "dangling trees" for the purpose of computing the CCE. (Note that this removal may change the counting numbers of regions connecting the core to the dangling trees.) To see this, follow an elimination order from the tree's leaf regions to the core. Since it is just message passing on the (hyper) tree and since the tree was already at a fixed point of GBP, the marginalization result will be equivalent to removing those regions.

Analogous to the factor graph case, cluster-cumulants defined on a region subgraph with two or more disconnected components vanish. The proof is identical to the one given for Theorem 3, and so we simply state the result:

THEOREM 6. *If a region subgraph $G_\alpha$ consisting of regions in the CRG that are descendants of $\alpha$ and that have counting numbers $c_\beta^\alpha \neq 0$ is disconnected, then the cluster-cumulant satisfies $C_\alpha = 0$.*

This is illustrated in Figure 3-Right. Note again that the partial log partition function may not vanish; rather, it will equal the sum of two (or more) terms whose contributions are already included in the lower order cumulants.

In building our CCE, it is clear that we should avoid CCs that correspond to either disconnected or singly connected region subgraphs. (However, if we happen to include them, no harm is done to the approximation; their contribution will be zero). More generally, Theorem 4 remains valid for region graphs, suggesting to define CCs over tightly connected groups of regions that are expected to exhibit strong dependencies beyond what are already modeled by the calibrated region graph. This still leaves considerable freedom to design a good CCE. The experiments in Section 5 will provide further guidance in his respect.

We end this section with a different perspective of what is accomplished with the CCE. Starting with Eqn. (5) we can write

$$\log Z_\psi = \log \hat{Z}_{BP} + \log Z_b \qquad (17)$$
$$= -\sum_{\alpha \in \mathcal{R}_{\text{calibrated}}} c_\alpha \mathcal{F}_\alpha - \sum_{\alpha' \in \mathcal{R}_{\text{uncalibrated}}} \kappa_{\alpha'} F_{\alpha'}$$

where $\mathcal{R}_{\text{calibrated}}$ and $\mathcal{R}_{\text{uncalibrated}}$ are calibrated and uncalibrated regions and $F_\alpha = -\log Z_\alpha$. The first term is the standard decomposition of the variational free energy for region graphs (see e.g. [11]) and the second term is the cluster cumulant expansion. This expression highlights the fact that CCE can be considered a correction to the variational free energy. Moreover, it suggests a procedure where increasingly many regions are moved from the uncalibrated part of the region graph to the calibrated part of the region graph. We leave exploration of these ideas for the future.

## 5 Experiments

We conducted a variety of experiments to study the CCE. Since the CCE is defined on a collection of clusters, we first describe the cluster choices used in our experiments. This description primarily refers to clusters that are sets of factors, i.e., for CCE on a factor graph. CCE on a region graph considers clusters that are sets of calibrated regions. In the sequel a $k$-cluster is a collection of $k$ factors (or regions).

We considered two different collections of clusters in our experiments. We first considered the poset $\Omega_{all}$ containing all pairs of factors, all triplets of factors, and so forth. Enumerating factors in this manner quickly becomes unmanageable, so the expansion must be truncated. We will use $\Omega_{all}^l$ to denote truncating the poset at level $l$, where for example $\Omega_{all}^4$ denotes the series truncated after all quintuplets of factors have been included.

The second collection of factors considered come from the Truncated Loop Series (TLS) algorithm of [2]. The TLS algorithm finds a subset of all generalized loops in a factor graph. It does so by first finding a set of $S$ simple loops (i.e., cycles in the factor graph with degree 2) and then *merges* these simple loops to create a set of generalized loops (i.e., cycles with degree $\geq 2$). In the TLS algorithm, a generalized loop is formed from two simple loops $l$ and $l'$ by finding a path in the factor graph from some factor or variable in $l$ to some factor or variable in $l'$. Since many paths may connect two simple loops, the set of generalized loops is restricted to paths of at most length $M$.

The authors of [2] provide code that enumerates a set of generalized loops $l_1,...,l_N$ and computes the LS approximation to $\log Z$ after every loop. In [2], the set of loops are placed in descending order by the magnitude of their contributions, $|U_l|$ (see 2.3 for details). To make the TLS algorithm an "anytime" algorithm, we do not post-process the set of loops and instead report the LS approximation on the order in which the loops are discovered. A sequence $\alpha_1,...,\alpha_N$ of clusters can be constructed from a sequence of loops $l_1,..,l_N$, where $\alpha_i$ is the set of factors in generalized loop $l_i$ [2]. In the experiments that follow 'LS (TLS)' denotes the Loop Series approximation on the (unsorted) sequence of loops from the TLS algorithm and 'CCE (TLS)' is the CC approximation on the unsorted sequence of loops.

---

[2] Since many generalized loops are defined over the same set of factors, we consider only the unique sets of factors.

When adding clusters to the CCE, it is important to note that the collection of clusters may become imbalanced in that the approximation in Eqn. (14) does not include all intersections (and intersections of intersections etc.) of all clusters, which leads to over-counting in the CCE. While adding clusters bottom up along the poset $\Omega_{all}$ guarantees a balanced CCE, this is not always true for the TLS sequence leading to suboptimal results. We thus emphasize that we include CCE (TLS) results for the sake of comparison, but that it is not the expansion that we recommend for the CCE.

We ran experiments on synthetic Markov Network (MN) instances as well as benchmark instances from the UAI-2008 solver competition. In all experiments, we take the absolute difference of true and approximate log-partition functions $|\log Z - \log \hat{Z}|$ as our error measure. We stop BP and GBP when $L_\infty(b_p(\mathbf{x}_c), b_c(\mathbf{x}_c)) < 1e^{-8}$, where $b_c(\mathbf{x}_c)$ is the belief at a child region $c$, $b_p(\mathbf{x}_c)$ is the marginal belief of parent region $p$ on variables $\mathbf{x}_c$ and $L_\infty$ is the maximum absolute difference between the two beliefs.

### 5.1 Grids

We first tested pairwise MNs defined on $10 \times 10$ grids. Each grid instance has unary potentials of the form $f_i(x_i) = [\exp(h_i); \exp(-h_i)]$ and pairwise potentials of the form:

$$f_{ij}(x_i, x_j) = \left[ \begin{array}{cc} \exp(w_{ij}) & \exp(-w_{ij}) \\ \exp(-w_{ij}) & \exp(w_{ij}) \end{array} \right]$$

The values of $h_i$ and $w_{ij}$ were drawn from $\mathcal{N}(0, \sigma_i^2)$ and $\mathcal{N}(0, \sigma_{ij}^2)$ distributions, respectively. To study the behavior of the CCE at various interaction strengths, we fixed $\sigma_i = 0.1$ and varied $\sigma_{ij}$ from 0.1 to 1. Reported errors are averages across 25 instances at each setting, and error bars indicate the standard error.

Figure 4 compares the different series approximations on grids. Following [2], the TLS algorithm was run on the $10 \times 10$ grid instances with $S = 1000$ and $M = 10$.

'CCE (BP)' is the CC approximation given a sequence of terms efficiently enumerated in $\Omega_{all}^{15}$. All pairs and triplets of factors on a pairwise grid have zero contribution and can be ignored. The only 4-clusters with non-zero contribution are the 81 faces of the $10 \times 10$ grid. To fill out the remaining levels of $\Omega_{all}^{15}$, we enumerate pairs, triplets and quadruplets of connected faces (sharing a vertex or edge), since the CC for two disconnected faces in the grid is zero.

'CCE (GBP)' is the CCE on a region graph. GBP was run on a region graph with outer regions equal to the faces of the grid. In this case, all pairs and triplets of faces give zero contributions and can be ignored. Each $3 \times 3$ subgraph in the grid is comprised of 4 faces forming a cycle. We take all

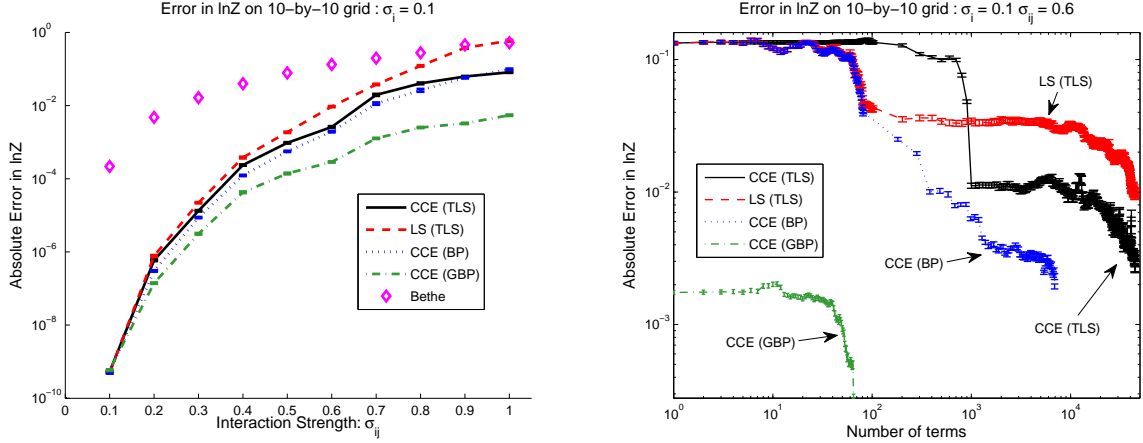

Figure 4: **Left:** Comparison of the 4 series approximations on $10 \times 10$ grids at a variety of interaction strengths. **Right:** $Error_Z$ versus the number of terms in each approximation, for $\sigma_{ij} = 0.6$.

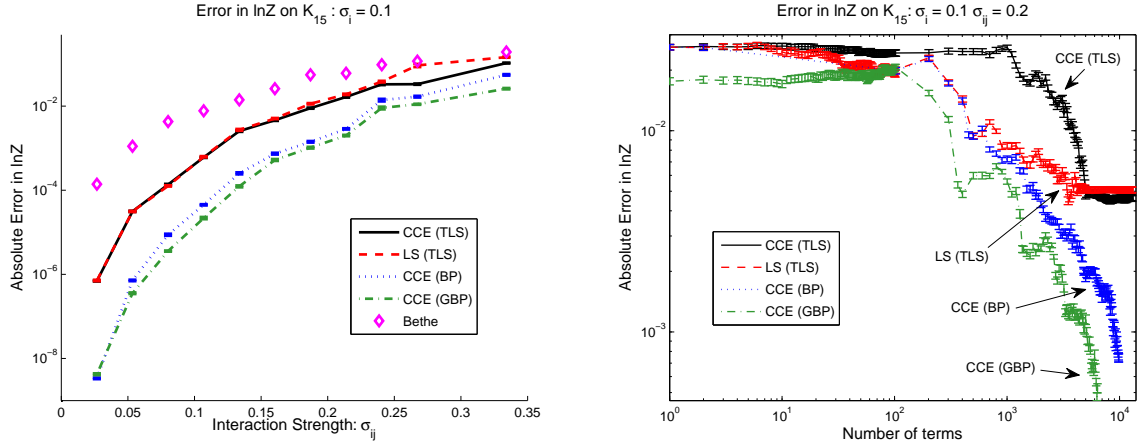

Figure 5: **Left:** Comparison of the 4 series approximations on complete MNs of 15 variables at different coupling levels. **Right:** $Error_Z$ as a function of the number of terms in each series.

64 of these to be the clusters in the 'CCE (GBP)' approximation. Importantly, the intersection of any two such clusters will be a cluster on either a disconnected or acyclic subgraph (having a zero contribution) implying that the whole collection of clusters remains balanced.

Figure 4-Left compares the series approximations at a variety of coupling levels. The error reported is w.r.t. the final $\log Z$ approximation of each series. All the approximations offer a substantial error reduction over the Bethe approximation for instances with weak interactions. As interaction strength is increased, the relative improvement of each series declines. At the strongest coupling level, the error in the LS approaches the Bethe approximation error, while CCE remains an order of magnitude better.

Figure 4-Right shows a trace of the error as a function of the number of terms included. Each LS term can be computed very efficiently for binary MNs, while each CCE term requires inference on a set of factors (or regions). Thus, comparing the series on the the number of terms may seem to unfairly favor the CC approximations. However, the LS requires enumerating generalized loops in the MN which is more expensive than enumerating clusters.

### 5.2 Complete Graphs

We also experimented on pairwise MNs over a complete graph of 15 variables. Here 'CCE (BP)' is the CC approximation given an efficient enumeration of terms in $\Omega_{all}^{14}$. Since all pairs of factors are acyclic (and have zero contribution), we begin by considering all cycles of length 3, which is equivalent to all embedded $K_3$ subgraphs. We then proceed with adding all $K_4$, all $K_5$ and finally all $K_6$ subgraphs to the CCE. Note that this enumeration is efficient in that it skips certain contributions, such as cycles over 4 variables because their contribution will be included when we add the corresponding $K_4$ subgraph.

'CCE (GBP)' enumerates in a similar fashion. GBP is run on a region graph formed using the "star" construction – outer regions equal to all cycles of length 3 passing through vertex 0. We then enumerate all complete graphs $K_4$, $K_5$ and $K_6$ containing vertex 0. The set of $K_4$ containing ver-tex 0 is only a subset of all the $K_4$'s embedded in the MN. However, enumerating embedded complete graphs in this fashion ensures that the collection of clusters remains balanced.

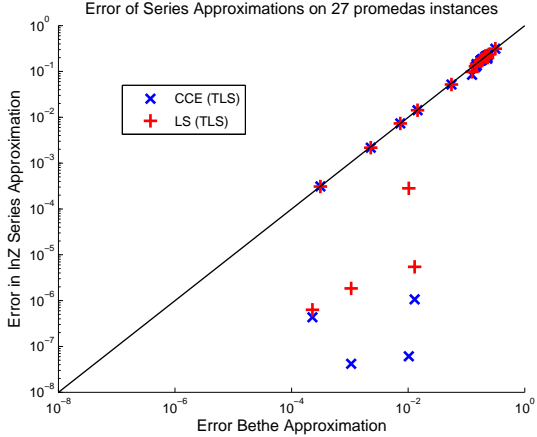
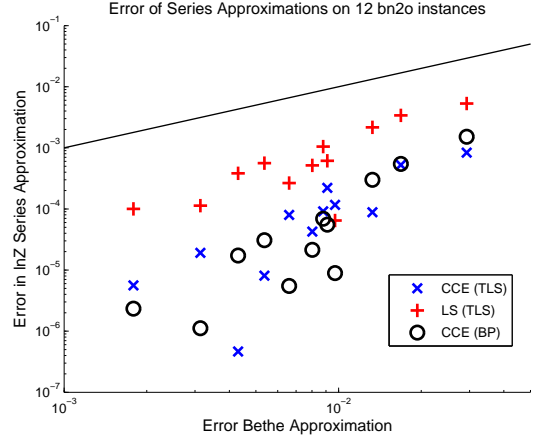

Figure 6: Scatter plot of $Error_Z$ for the LS and CC approximations versus error in the Bethe approximation. Points below the line are improvements upon IBP.

Figure 7: Scatter plot of $Error_Z$ for each series approximation versus error in the Bethe approximation. Points below the line are improvements upon IBP.

Figure 5 compares the four series approximations across a variety of interaction strengths. In these experiments the TLS algorithm was run with $S = 5000$ and $M = 10$. On complete graphs, the CC and LS approximations behave similarly when using terms from the TLS algorithm. The CCE as described above is much more accurate. This is because while the TLS algorithm enumerates $> 10K$ loops, the loops contain at most 5 factors while the CCE up to $K_6$ covers 15 factors.

### 5.3 UAI-2008 Benchmarks

In addition to the synthetic instances, we evaluated each of the series approximations on instances from the 2008-UAI solver competition. We selected 12 bn2o instances and 27 promedas instances that were solvable by our Junction Tree implementation. The promedas instances have a very sparse graph structure and contain between 400-1000 factors. As a result, enumerating all clusters in $\Omega_{all}^l$ is costly and yields few non-zero terms even for small $l$. Thus, for the promedas instances we only compare the LS and CC approximations on the loops found by the TLS algorithm. Figure 6 shows the error of the CC and LS approximations versus the Bethe approximation. Points below the diagonal line indicate improvement over the Bethe approximation. The TLS algorithm was run on all 27 instances with $S = 100$ and $M = 10$. These results are consistent with [2] in that if the TLS algorithm finds most of the generalized loops in an instance, the error is reduced by several orders of magnitude, while if a small set of all generalized loops are found, the series approximations offer no improvement.

Unlike the promedas instances, the bn2o instances have densely connected graph structures and contain far fewer factors. These instance are thus better suited to enumeration along $\Omega_{all}$. Figure 7 compares the error of the 4 series approximations to the error of the Bethe approximation. The TLS algorithm was run with $S = 10K$ and $M = 10$, producing more than 10K generalized loops. 'CCE (BP)' was run on the set of clusters in $\Omega_{all}^4$.

## 6 Conclusion

We have introduced a new cluster-cumulant expansion based on the fixed points of either BP or GBP. The expansion was inspired by the LS of [1] but has certain advantages over the latter. First, terms corresponding to disconnected clusters vanish. More generally, only tightly coupled groups of variables are expected to make significant contributions, which can be used to significantly cut down on the number clusters that need to be considered. Second, while the LS was only developed for binary variables, the CCE is defined on arbitrary alphabets. Third, the CCE has a natural extension to GBP on region graphs. Finally, the accuracy of the CCE expansion improves upon the LS.

The CCE represents a very natural extension of the Kikuchi approximation as it is based on the same type of expansion. But unlike GBP on a region graph, the CCE does not suffer from convergence issues and does not require storing beliefs during message passing. This makes it a useful "anytime" tool to improve results obtained form GBP. It also suggests new algorithms that move regions from the CCE to regions used in GBP. The question of which regions should be included in the GBP + CCE approximation and whether a region should be included in GBP or be handled by the CCE are left for future investigation.

### Acknowledgements

MW acknowledges support by NSF Grants No. 0914783, 0928427, 1018433. AI acknowledges support by NSF IIS grant No. 1065618.